%% file: root.tex
\documentclass[a4paper, 10pt, conference]{ieeeconf}      % Use this line for a4 paper

\IEEEoverridecommandlockouts                              % This command is only needed if 
\DeclareMathAlphabet{\mathcal}{OMS}{cmsy}{m}{n}
\DeclareSymbolFont{largesymbols}{OMX}{cmex}{m}{n}

\usepackage{graphics} % for pdf, bitmapped graphics files
\usepackage{epsfig} % for postscript graphics files
\usepackage{mathptmx} % assumes new font selection scheme installed
\usepackage{times} % assumes new font selection scheme installed
\usepackage{amsmath} % assumes amsmath package installed
\usepackage{amssymb}  % assumes amsmath package installed
\usepackage{color}
\usepackage{bm}
\usepackage{subfigure}
\usepackage{subcaption}
\usepackage{float}
\usepackage{graphicx}
\usepackage{multirow}
\usepackage{tabularx}

\title{\LARGE \bf
Intelligent Spatial Perception by Building Hierarchical \\ 3D Scene Graphs for Indoor Scenarios with the Help of LLMs$^{*}$
}

\author{Yao Cheng$^{1,2}$, Zhe Han$^{2}$, Fengyang Jiang$^{1,2}$, Huaizhen Wang$^{1,2}$, Fengyu Zhou$^{3}$, 
Qingshan Yin$^{2}$, and Lei Wei$^{2}$
\thanks{*This research was funded by Key R\&D Program of Shandong Province, China of grant number 2023CXPT094 and the Jinan City and University Cooperation Development Strategy Project under Grant JNSX2023012.}% <-this % stops a space
\thanks{$^{1}$Shandong New Generation Information Industrial Technology Research Institute, Jinan, 250100, Shandong, P. R. China
        {\tt\small {chengyao01, jiangfy, wanghuaizhen@inspur.com}}}%
\thanks{$^{2}$Inspur Intelligent Terminal Co., Ltd., Jinan, 250100, Shandong, P. R. China
        {\tt\small {hanzhe, yinqsh, weilei03@inspur.com}}}%
\thanks{$^{3}$School of Control Science and Engineering, Shandong University, Jinan, 250061, Shandong, P. R. China
        {\tt\small zhoufengyu@sdu.edu.cn}}%
}

\begin{document}

\maketitle
\thispagestyle{empty}
\pagestyle{empty}

%%%%%%%%%%%%%%%%%%%%%%%%%%%%%%%%%%%%%%%%%%%%%%%%%%%%%%%%%%%%%%%%%%%%%%%%%%%%%%%%
\begin{abstract}
%This paper presents a novel approach to enhancing intelligent robot navigation through the development of hierarchical 3D Scene Graphs (3DSGs) enriched by Large Language Models (LLMs). 
%By integrating geometric and semantic information, our system provides a comprehensive spatial understanding, facilitating more effective navigation and task execution in complex environments.

%Intelligent robot navigation is a complex task that requires a deep understanding of the environment. While Large Language Models (LLMs) have shown promise in enhancing navigation capabilities, they require more than just geometric information to operate effectively. Traditional maps that contain only geometric data are insufficient for providing the contextual understanding necessary for intelligent navigation. The motivation for this research is to develop a system that can create smart spatial perceptions by building hierarchical 3D Semantic Graphs (3DSGs) with the assistance of LLMs, thereby enabling more sophisticated and context-aware navigation.
This paper addresses the high demand in advanced intelligent robot navigation for a more holistic understanding of spatial environments,  
by introducing a novel system that harnesses the capabilities of Large Language Models (LLMs) 
to construct hierarchical 3D Scene Graphs (3DSGs) for indoor scenarios. 
%, thereby enhancing spatial perception. 
The proposed framework constructs 3DSGs consisting of a fundamental layer with rich metric-semantic information,  
an object layer featuring precise point-cloud representation of object nodes as well as visual descriptors, 
and higher layers of room, floor, and building nodes.  
%and higher layers representing areas or regions, culminating in scalable representations through room, floor, and building nodes. 
Thanks to the innovative application of LLMs, 
not only object nodes but also nodes of higher layers, e.g., room nodes, are annotated 
in an intelligent and accurate manner. 
%in generating node attributes and summarizing descriptions is pivotal for context-aware navigation and task planning. 
A polling mechanism for room classification using LLMs is proposed to 
enhance the accuracy and reliability of the room node annotation. 
Thorough numerical experiments demonstrate the system's ability to integrate semantic descriptions with geometric data, creating an accurate and comprehensive representation of the environment instrumental for context-aware navigation and task planning. 
%This paper introduces a novel system that harnesses the capabilities of Large Language Models (LLMs) to construct hierarchical 3D Scene Graphs (3DSGs) for indoor scenarios, thereby enhancing spatial perception. The proposed method integrates a fundamental layer with metric-semantic information, an object layer with visual descriptors, and higher layers representing areas or regions, culminating in scalable representations through room, floor, and building nodes. The innovative application of LLMs in generating node attributes and summarizing descriptions is pivotal for context-aware navigation and task planning. A polling mechanism for room classification using LLMs is proposed to improve the accuracy and reliability of room node annotation. Thorough experiments demonstrate the system's ability to integrate semantic descriptions with geometric data, creating an accurate and comprehensive representation of the environment instrumental for intelligent agent navigation and task planning.

\end{abstract}

%%%%%%%%%%%%%%%%%%%%%%%%%%%%%%%%%%%%%%%%%%%%%%%%%%%%%%%%%%%%%%%%%%%%%%%%%%%%%%%%
\section{INTRODUCTION}
\input{intro.tex}
\section{METHOD}
\input{main.tex}

\section{EXPERIMENTAL RESULTS}
\label{sec:res}
\input{res.tex}
\section{CONCLUSION}
%The proposed system demonstrates the potential of combining geometric and semantic information through hierarchical 3DSGs and LLMs to create a smart spatial perception system. Future work will focus on refining the polling mechanisms and expanding the semantic descriptors to include dynamic environmental changes.
This research presents a significant advancement in the field of intelligent spatial perception for robotics by leveraging the power of LLMs to construct hierarchical 3DSGs. The proposed system has been validated through extensive experiments, demonstrating its effectiveness in generating accurate and comprehensive spatial representations for indoor environments. The innovative use of LLMs for node attribute generation and room classification has been shown to enhance the semantic richness and scalability of 3DSGs, which are crucial for intelligent navigation and task planning. The polling mechanism introduced for room classification improves annotation accuracy and addresses the challenges of multi-functional room perception. The findings from this study lay a solid foundation for future research in 3DSG construction and the application of LLMs in robotics, paving the way for more sophisticated and context-aware intelligent agents.
\bibliography{reference}

\end{document}

%% file: intro.tex
%The evolution of intelligent robot navigation calls for a significant shift in paradigm. It is no longer sufficient to rely solely on geometric mapping; instead, a more profound and holistic comprehension of spatial environments is imperative. The prevalent mapping techniques, as referenced in [1] and [2], are limited to geometric data, failing to encapsulate the semantic richness essential for true intelligence in navigation.

%3D Scene Graphs (3DSGs) have emerged as a robust representational framework for spatial environments. They offer a structured and comprehensive grasp of both geometric and semantic facets of a scene, as articulated in [3], [4], [5], and [6]. The utility of 3DSGs extends to various robotic applications, including navigation, object manipulation, and scene interpretation, as discussed in [7], [8], and [9]. Pioneering works such as those by Kim et al. [1] and Armeni et al. [2] have laid the conceptual groundwork for 3DSGs, emphasizing their significance in artificial intelligence and robotics for advanced tasks.

The advancement of intelligent robot navigation calls for a paradigm shift from mere geometric mapping 
to a more comprehensive understanding of spatial environments. 
Current widely adopted mapping techniques~\cite{MT17},~\cite{Labb2018RTABMapAA},~\cite{JC24} 
rely solely on geometric information 
and thus fail to capture the rich semantic context required for truly intelligent navigation. 
3D Scene Graphs (3DSGs)~\cite{KPS19},~\cite{Armeni20193DSG},~\cite{conceptgraphs},~\cite{HCH24} 
have emerged as a powerful representation of spatial environments 
that provide a structured and comprehensive understanding 
of both the geometric and semantic aspects of a scene. 
Therefore, they find applications in the field of robotics for tasks such as navigation, 
object manipulation, and scene interpretation
~\cite{Agia2022TaskographyER},~\cite{conceptgraphs},~\cite{Rana2023SayPlanGL}, \cite{RPH22}.  
%One of the key advantages of 3DSGs is their compatibility with natural language processing. 
%The attributes and relationships within the graph can be described using natural language, making them easily parsable by Large Language Models (LLMs). 
%This feature is particularly important as it allows for seamless integration with advanced AI systems that can interpret and generate human-like text.

%The increasing capabilities of LLMs~\cite{openai2024gpt4} have significantly impacted the utility of 3DSGs. With the improvement in handling larger tokens, LLMs can now process more complex and detailed 3DSG representations. This advancement has relieved previous restrictions on the size and complexity of the scenes that can be effectively understood and reasoned about by AI systems.

%The integration of Large Language Models (LLMs) into this domain has the potential to revolutionize how robots perceive and interact with their surroundings. 
%The literature on intelligent robot navigation has seen a significant shift towards incorporating semantic understanding into spatial representations. 
Early works, such as~\cite{KPS19},~\cite{Armeni20193DSG}, 
presented the first 3DSG construction frameworks and 
validated the idea of generating a graph spanning an entire building and 
incorporating rooms as well as objects detected together with the spatial relationships of these entities. 
%introduced the concept of 3D Scene Graphs (3DSGs) 
%as a powerful representation of spatial environments, providing a structured and comprehensive understanding of both the physical and semantic aspects of a scene.
The concept of spatial perception for robotics was first introduced in~\cite{HCH24}, 
where the significance of a hierarchical structure for a 3DSG was addressed 
and an advanced spatial perception system, as a collection of 
algorithms designed to generate 3DSGs with sensor data in real-time was developed~\cite{HCH24}. 
These works have laid the conceptual foundation for subsequent research in 3DSGs, 
emphasizing their significance in Artificial Intelligence (AI) and the field of robotics.

The advent of Large Language Models (LLMs)~\cite{openai2024gpt4},~\cite{Touvron2023LLaMAOA} and 
Large Visual Language Models (LVLMs)~\cite{NEURIPS2023_6dcf277e},~\cite{2021Learning} 
has been a catalyst in advancing the capabilities of 3DSGs. 
The synergy between LLMs and spatial understanding holds the potential 
to transform the way robots perceiving and interacting with their environments. 
Recent studies~\cite{conceptgraphs} have begun to explore the fusion of LLMs with 3DSGs, 
demonstrating the potential for enhanced task planning and navigation.
The ability to describe node attributes and relationships within the graph using natural language is a pivotal advancement, 
facilitating seamless integration with AI systems that can generate and interpret human-like text. 
However, the authors~\cite{conceptgraphs} only address the construction of an object-centric single-layer 3DSG, 
leaving the advantages of LLMs in establishing scalable hierarchical 3DSGs unexploited. 

Building upon this rich body of literature, 
this paper introduces a novel system that 
leverages the power of LLMs to construct hierarchical 3DSGs
for intelligent spatial perception in indoor scenarios. 
%We propose a multi-tiered structure of 3DSGs 
%that captures the topological organization of spatial relationships at various levels of abstraction, 
%ranging from individual objects to entire rooms or buildings. 
We outline a method for constructing 3DSGs 
that encompass a fundamental layer with metric-semantic information, 
an object layer with visual descriptors as well as point-cloud representations, 
and higher layers representing areas or regions, 
culminating in scalable representations through room, floor, and building nodes. 
LLMs are leveraged to annotate not only object nodes but also nodes of higher layers, e.g., room nodes. 
%We demonstrate the innovative application of LLMs in generating node attributes 
%and summarizing descriptions, which is pivotal for context-aware navigation and task planning. 
In particular, a polling mechanism for room classification using LLMs is proposed 
to improve the accuracy and reliability of the room node annotation. 
As a byproduct, a nice problem formation of constructing 3DSGs and exploiting LLMs as a tool 
is established for future research in this area. 
Thorough experiments show that, by harnessing the capabilities of LLMs, 
our system is able to integrate semantic descriptions with geometric data and 
to create an accurate and comprehensive representation of an indoor environment 
that is instrumental for intelligent agent navigation and task planning.

%% file: main.tex
\subsection{Problem formulation} 
\begin{figure*}[htbp]
%         \vspace{-0.8cm}   %调整图片与上文的垂直距离  
%	\setlength{\abovecaptionskip}{0.cm} %调整标题上方的距离   
%	\setlength{\abovecaptionskip}{0.1cm} %调整标题下方的距离 	   
   % \ContinuedFloat
    \centering
%    \subfigure[Illustration of the indoor tour-guide robot (left) and 
%    the indoor delivery robot (right)]{
        \includegraphics[width=0.8\textwidth]{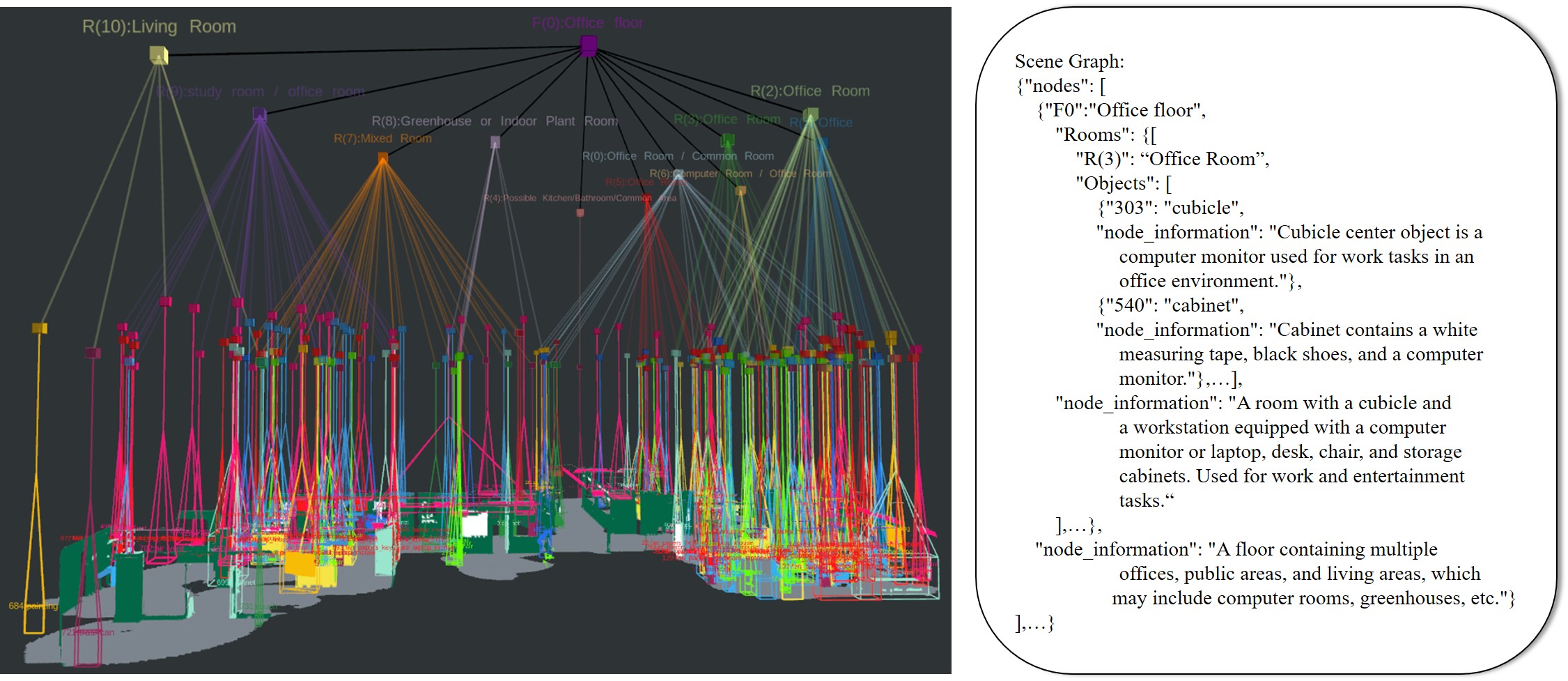}
%        }
        \caption{Illustration of the proposed hierarchical 3DSG}
    \label{fig:3DSG_illu}
\end{figure*}
%with math notations
%3DSGs are characterized by their topological organization, which captures the spatial relationships between objects in an environment. 
%This includes relative positions, orientations, and distances. 
The hierarchical nature of 3DSGs allows for the representation of complex scenes at various levels of abstraction, 
from individual objects to entire rooms or buildings~\cite{Armeni20193DSG}, \cite{HCH24}. 
Here we denote a 3DSG consisting of $K$ hierarchical layers as 
$\mathcal{G} = (\mathcal{V}, \mathcal{E})$~\cite{KPS19}, \cite{Armeni20193DSG}, \cite{HCH24}, 
where $\mathcal{V}$ represents the set of nodes belonging to the $K$ layers, 
i.e., $\mathcal{V} = \cup_{k = 1}^{K} \mathcal{V}_k$ 
with $\mathcal{V}_k$ containing the $M_k$ nodes of the $k$-th layer 
\begin{equation}
\mathcal{V}_k = \left\{v_{k,1}, v_{k,2}, \ldots, v_{k,M_k}\right\}, 
\end{equation}and $v_{k,j}$ denoting the $j$-th node ($j = 1, 2, \ldots, M_k$) 
of the $k$-th layer. 
Each node is annotated by a set of descriptions, i.e., for node $v_{k,j}$ 
\begin{equation}
\mathcal{C}_{k,j} = \left\{c_{k,j}^{(1)}, c_{k,j}^{(2)}, \ldots, c_{k,j}^{(L_{k,j})}\right\}, 
\end{equation}consisting of $L_{k,j}$ attributes covering multiple perspectives and dimensions. 
These pieces of information serve as the basis for intelligent agent navigation and task planning. 
For instance, the attributes of object nodes include descriptions that 
distinguish ``operable'' objects including various small items such as cups and food 
from assets like furniture and home appliances 
that an intelligent agent generally cannot move. 
%Different from "room" and "building" nodes, 
%Moreover, the description of an object node $v_{k,j}$ 
%also contains its point-cloud presentation $\bm{P}_{k,j}^{\rm{o}}$ and 
%a semantic feature vector $\bm{f}_{k,j}^{\rm{o}}$ obtained 
%based on CLIP (Contrastive Language-Image Pretraining)~\cite{2021Learning}.
%which is used for object tracking.  
%{\color{red}{which are used for the representation of the geometric position of the node and the identification and comparison of objects.}}
%DINO (self-Distillation with NO labels)

The whole set of edges in the 3DSG $\mathcal{G}$ is represented by $\mathcal{E}$.  
Each edge connects two nodes within the same layer or two nodes belonging to adjacent layers, i.e., 
an edge stemming from node $v_{k,j}$ of the $k$-th layer connects only nodes in 
$\mathcal{V}_{k - 1} \cup \mathcal{V}_k \cup \mathcal{V}_{k + 1}$~\cite{HCH24}.

\subsection{Construction of 3DSGs} 
\label{subsec:3DSG}
This section delves into the generation of hierarchical 3DSGs with the help of LLMs. 
Focusing on indoor scenarios, we propose a multi-tier structure of the 3DSG 
illustrated in {Figure~\ref{fig:3DSG_illu}}. 
We call {\textit{Layer-1}} a fundamental layer taking the form of a 3D mesh 
and holding crucial metric-semantic information 
required for the construction of upper layers of the 3DSG and navigation tasks. 
In parallel to building {\textit{Layer-1}}, object nodes are initiated 
and gradually equipped with rich geometric as well as semantic descriptions,  
forming {\textit{Layer-2}}. 
In {\textit{Layer-3}}, object nodes are grouped, 
populating the concept of areas or regions of the scenario that the 3DSG is describing. 
By resorting to LLMs, the resulting areas are annotated 
based on the attributes of the object nodes they contain. 
 {\textit{Layer-4}} and {\textit{Layer-5}} are designed to hold ``floor'' and ``building'' nodes, respectively, 
 which are particularly useful for generating a scalable and well-organized representation 
 of, e.g., large multi-functional venues. 
%with the sparse graph obtained by approximating the Generalized Voronoi
%Diagram (GVD) which is computed based on the 3D mesh of {\textit{Layer-1}}
%including a fundamental layer, an object layer, a room layer, a floor layer, and a building layer. 
%\begin{itemize}
%a fundamental layer, an object layer, a room layer, a floor layer, and a building layer. 
%\end{itemize}

\subsubsection{Layer-1 - Fundamental layer}
\label{subsubsec:layer1}
As the first layer of the 3DSG, we construct a metric-semantic 3D mesh. 
%utilizing Kimera and Voxblox.
A visual-inertial~\cite{RA20} or a LiDAR-visual-inertial odometry pipeline~\cite{Labb2018RTABMapAA},~\cite{JC24},  
is available as a common practice.  
For each RGB-D frame\footnote{Depending on the type of cameras used, the depth images could be directly provided 
by an RGB-D camera or retrieved 
through the stereo matching in the case of a stereo camera.} 
coming in, we first employ the 2D semantic segmentation network\footnote{
Class-agnostic segmentation models can also be employed to 
achieve open-vocabulary detection and segmentation~\cite{conceptgraphs}.
}
 YOLOv8~\cite{Yolov8}
to achieve a pixel-level semantic segmentation of the RGB image. 
This semantic segmentation and depth information are then converted into a semantically annotated 3D point cloud, 
which is adjusted based on the robot's pose estimate, 
i.e., it is transformed from the coordinate system of the camera to 
the world coordinate system. 
%Specifically, we sustain a voxel-based, metric-semantic map that is dynamically centered on the robot's vicinity. 
On the one hand, this semantically labeled point cloud is integrated into 
a Truncated Signed Distance Field (TSDF) via a ray-casting process~\cite{HCH24}. 
The resulting voxel-based map is equipped with data indicating the availability of free space 
and a probabilistic distribution of potential semantic labels for each voxel~\cite{RA20}. 
Via the marching cubes algorithm, 
the voxel map is transformed into a 3D mesh~\cite{HCH24}. 
On the other hand, the aforementioned point cloud is used to 
construct object nodes of {\textit{Layer-2}} 
as discussed in detail in Section~\ref{subsubsec:layer2}. 

%This approach with an active window eliminates the necessity of maintaining a large and memory-intensive voxel-based representation of the entire environment, as was initially proposed in [34]. Instead, it enables a progressive conversion of the voxel-based map within the active window (which moves with the robot) into a more streamlined 3D mesh.

%These algorithms generate a localized Euclidean Signed Distance Function (ESDF) 
%centered on the robot's current position.
In addition to the 3D mesh used also for the visualization purpose, 
{\textit{Layer-1}} contains a ``hidden'' sub-layer 
which plays an essential role in building the room layer 
described in Section~\ref{subsubsec:layer3}. 
It takes the form of a sparse graph of places as first introduced in~\cite{HCH24}, 
where a Generalized Voronoi Diagram (GVD) of the environment 
is computed based on the 3D mesh and approximated 
to form the graph of places as a representation of 
obstacle-free spots and their connectivity. 

%The algorithm includes the following steps:
%
%1. Layer Initialization: Initialize the base layer with geometric nodes representing the spatial structure.
%
%2. Semantic Enrichment: Utilize LLMs to add semantic attributes to each node in $\mathcal{V}$.
%
%3. Edge Construction: Define edges $\mathcal{E}$ based on spatial relationships and semantic connections between nodes.
%
%4. Hierarchical Integration: Integrate layers to form a hierarchical structure that captures different levels of abstraction.
%
%\begin{equation}
%\text{Layer} = f(\text{Geometric Nodes}, \text{Semantic Attributes}, \text{Spatial Relationships})
%\end{equation}

%A Table for the Algorithm

%This work outlines the foundational aspects of our research, providing a roadmap for the development and implementation of smart spatial perception systems. The following sections will delve into the specific experimental designs, results, and analyses that support our claims.
  
\subsubsection{Layer-2 - Object layer}
\label{subsubsec:layer2}

%Starting with a collection of RGB-D images that include poses, we employ a class-agnostic segmentation model to identify potential objects. 
In Section~\ref{subsubsec:layer1}, where the construction of the fundamental layer is detailed, 
the point cloud retrieved from each entire data frame is used to build up a global map to 
thoroughly represent the spatial structure of the environment. 
For {\textit{Layer-2}}, nevertheless, the point cloud is extracted on 
the ``segmentation-mask'' basis and used to initiate a new object node 
or to improve the representation of an existing one. 
The corresponding mask serves as the input of 
a visual feature extractor such as CLIP~\cite{2021Learning} to 
generate a visual descriptor. 
Consequently, the point cloud together with the visual descriptor 
are taken as the geometric and semantic representation of each object instance. 
By leveraging both geometric and semantic similarity metrics~\cite{conceptgraphs}, 
the association of these object instances is conducted.  
Finally, for an object node $v_{2,j}$ of {\textit{Layer-2}} ($k = 2$), 
the point cloud of its various instances is fused into its unique 
point-cloud representation $\bm{P}_{2,j}^{(\rm{o})}$, whereas 
a semantic feature vector $\bm{f}_{2,j}^{(\rm{o})}$ is obtained likewise 
through the object instance fusion~\cite{conceptgraphs}.

Subsequently, with the help of an LVLM and an LLM, 
node attributes are generated.  
%Subsequently, we apply a LVLM to 
%generate descriptions for each node 
%and utilize a LLM to deduce the connections 
%between adjacent nodes, thereby forming edges within the graph. 
%After the complete image sequence has been analyzed, a vision-language model, referred to as LVLM, is employed to create captions for objects. For each detected object, the model is fed the best 10 image crops from the most favorable views, along with the instruction "describe the central object in the image," to produce an initial set of rough captions
Based on the RGB image of each object instance, an LVLM is first used 
to produce an initial set of object descriptions covering the following aspects:
\begin{itemize}

\item
Object state: The current condition or status of an object (e.g., switched on or off, orientation).

\item
Predicates: Descriptive statements that relate objects and their attributes (e.g., ``the cup is on the table'').

\item
Affordances: Action possibilities presented by objects based on their properties (e.g., ``a chair can be sat on'').

\item
Other attributes: Detailed characteristics of objects that can influence task planning and execution.

\end{itemize}
The prompts for the LVLM are adapted 
according to observations and findings in navigation tasks where 3DSGs are used. 
Along with the object association and fusion, an LLM represented by 
$\phi_{\rm{LLM}}$ is prompted to 
provide a summary of the descriptions associated with the object instances belonging to the same object node, 
i.e., 
\begin{equation}
\mathcal{C}_{2,j} = \phi_{\rm{LLM}}\left(\mathcal{C}_{2,j}^{({\rm{LVLM}},1)}, \ldots, \mathcal{C}_{2,j}^{({\rm{LVLM}},N_{2,j})}, P_{2,j}\right),
\end{equation}where $\mathcal{C}_{2,j}^{({\rm{LVLM}},i)}$ ($i = 1, \ldots, N_{2,j}$) 
denotes the node description obtained via the LVLM in the $i$-th instance 
of the $j$-th object node of {\textit{Layer-2}}, 
and $P_{2,j}$ represents the prompt tailored for the LLM 
to output a desired summary of the node description. 

Note that the procedure of node captioning described above 
can be carried out in real-time by calling the API of an LVLM and an LLM. 
In addition, we have an offline alternative~\cite{conceptgraphs} in our implementation as well  
based on the sequence of RGB images saved. 
It is particularly useful for cases where the LVLM and the LLM are deployed and run locally 
on the intelligent agent. 
Moreover, it can also be treated as a separate module 
used to supplement and enrich the node attributes of an exiting 3DSG~\cite{Armeni20193DSG} 
providing that the raw data is available or can be extracted from an environment~\cite{shenigibson}. 

%For instance, the attributes of object nodes include descriptions that 
%distinguish "operable" objects including various small items such as cups and food 
%from assets like furniture and home appliances 
%that an intelligent agent generally cannot move. 
%%Different from "room" and "building" nodes, 
%Moreover, the description of an object node $v_{k,j}$ 
%also contains its point-cloud presentation $\bm{P}_{k,j}^{\rm{o}}$ and 
%a semantic feature vector $\bm{f}_{k,j}^{\rm{o}}$ obtained 
%based on CLIP (Contrastive Language-Image Pretraining) or 
%DINO (self-Distillation with NO labels), 
%which is used for object tracking.  	

%Encoding Information for Task Planning

%- Geometric properties: Volume $V$, Surface Area $A$

%{\color{red}{- Semantic descriptors: Object function $F_o$, Room purpose $P_r$, derived through LLM analysis. + math formulation of using LLM and with prompts as input to get node description or node types}} 

Compared to the object layer of Hydra~\cite{HCH24}, 
the geometric representation of the object nodes in {\textit{Layer-2}} of our proposed 3DSG 
 is more accurate and thus more suitable for robotic manipulation tasks. 
In addition, each node is enriched with a semantic descriptor 
and a comprehensive set of attributes derived through LLM-based analysis, 
which are crucial for intelligent task planning.

%
%By encoding this information into the 3DSG 
%
%a variety of information necessary for task planning
	
\subsubsection{Room layer and upper layers}
\label{subsubsec:layer3}
Though intuitive and straightforward for human-beings, 
room identification is a non-trivial robotic perception task. 
To initiate the construction of the room layer, 
we segment the rooms based on the sub-layer of {\textit{Layer-1}} described in Section~\ref{subsubsec:layer1} 
by clustering this sparse graph of places using persistent homology~\cite{HCH24}. 
%derive a topological representation of spatial areas from the ESDF, 
%and subsequently categorize these areas into distinct rooms by employing methods reminiscent of community detection in network analysis.
Once the room identification is done, we propose to 
categorize and annotate the rooms by prompting an LLM to infer their
functional and structural characteristics
based on the object nodes of {\textit{Layer-2}}.  
In this way, the description set of the $\ell$-th room node of {\textit{Layer-3}} is obtained as 
\begin{equation}
\mathcal{C}_{3,\ell} = \phi_{\rm{LLM}}\left(\underset{j \in \mathcal{S}_{3, \ell}^{(\rm{o})}}{\cup}\mathcal{C}_{2,j}, P_{3,\ell}\right)
\end{equation}where $\mathcal{S}_{3, \ell}^{(\rm{o})}$ represents 
an index set of object nodes of {\textit{Layer-2}} belonging to the $\ell$-th room node $v_{3,\ell}$, 
and $P_{3,\ell}$ symbolizes the prompt template designed for the LLM 
to deduce a room type and perform room node captioning 
based on the object nodes that room node $v_{3,\ell}$ contains 
(cf. Figure~\ref{fig:prompt_template_ori} for an example 
where the LLM is guided to infer a room type). 

Annotating room nodes featuring typical functional characteristics 
with explicit labels facilitates down-streaming planning tasks~\cite{HCH24}. 
In addition, the token size can be reduced, and providing the LLM with 
concise and explicit information prevents it from hallucinating.  
%When it is possible, annotating a room node with an accurate label is important, 
%for the token size can be reduced and providing the LLM with 
%concise and explicit information prevents it from hallucinating effectively.  
Inaccurate node annotation, on the other hand, can be misleading, 
resulting in planning failures. 
Hence, we propose a polling mechanism that leverages 
the LLM's own capabilities to resolve issues with hallucination as well as uncertainty 
and to effectively identify multi-functional segments of an indoor environment 
that should be annotated with a concise description instead.
%and to enhance the credibility of the deduced room labels.
To this end, 
given a set of typical room labels of size $N$ 
\begin{equation}
\mathcal{C}^{(\rm{typ})} = \left\{c^{(\rm{label})}_1, c^{(\rm{label})}_2, \ldots, c^{(\rm{label})}_N\right\}, 
\end{equation}we query the LLM for $L$ rounds and obtain the polling results as follows 
%\begin{equation}
%p_{n} = {\sum}_{i = 1}^{L} 
%\phi_{\rm{LLM}}^{(i)}\left(\underset{j \in \mathcal{S}_{3, \ell}^{\rm{o}}}{\cup}\mathcal{C}_{2,j}, \mathcal{C}^{(\rm{typ})}, \hat{c}_{3,\ell}^{(\rm{label})}, P_{3,\ell}^{(\rm{label})} \right), 
%\end{equation}
\begin{equation}
\bm{p}_{\ell} = {\sum}_{i = 1}^{L} 
\phi_{\rm{LLM}}^{(i)}\left(\underset{j \in \mathcal{S}_{3, \ell}^{(\rm{o})}}{\cup}\mathcal{C}_{2,j}, \mathcal{C}^{(\rm{typ})}, P_{3,\ell}^{(\rm{label})} \right), 
\end{equation}where $\bm{p}_{\ell} = \left[\begin{array}{cccc} p_{\ell, 1} & p_{\ell, 2} & \ldots & p_{\ell, N} \end{array}\right]^{\rm{T}}$ 
with $p_{\ell, n}$ representing for how many times of the $L$ query rounds, the typical room label $c^{(\rm{label})}_n$ 
is selected, and $P_{3,\ell}^{(\rm{label})}$ is the corresponding prompt template. 
As discussed in detail in Section~\ref{subsec:res_room}, 
for indoor household scenarios, the set of typical room labels are 
determined empirically by evaluating 483 rooms of the Stanford 3D Scene Graph dataset~\cite{Armeni20193DSG}. 
Based on findings of a through investigation shown in Section~\ref{subsec:res_room}, 
we propose to annotate room node $v_{3,\ell}$ with 
$c^{(\rm{label})}_n$ only when $p_{\ell, n} = L$, 
i.e., for all $L$ rounds of querying the LLM, 
$c^{(\rm{label})}_n$ is suggested as the room label for $v_{3,\ell}$. 
This strategy guarantees the accuracy of 
the room labels incorporated in the 3DSG to a large extent 
and in the meantime effectively identifies multi-functional segments 
of the scenario, where a description provides 
more insights into down-streaming navigation tasks 
compared to a single label. 
Moreover, 
sometimes the identified room areas might not 
have stand-out or widely acknowledged functionality characteristics, 
at least not according to the object nodes they contain, e.g., 
lobbies, corridors, closets. 
Therefore, for all these cases, a room label is not explicitly provided in the 3DSG, 
and instead a concise version of the room node description is included 
which is able to supply subsequent robotic tasks with useful and unambiguous information.

For the construction of upper layers of the 3DSG, 
taking the floor layer as an example, 
we cluster rooms with a similar height determined with the poses of the object nodes they contain 
and form a floor layer.  
Then the LLM is guided to annotate the resulting floor node 
based on the room nodes of this floor. 
In the case of multi-storey high-rise buildings, a robotic agent usually travels with an elevator to 
reach a certain floor, and the 3DSG of each floor 
might have to be constructed separately. 
Floor layers are naturally identified with a-priori knowledge of the environment. 

%The integration of hierarchical 3DSGs with the semantic enrichment provided by LLMs offers a robust framework for smart spatial perception. 
%This approach not only enhances the robot's understanding of its environment 
%but also paves the way for more sophisticated navigation algorithms that can adapt to various spatial contexts.		
%
%This system allows the robot to adapt its navigation strategy to the specific type of room it is navigating, 
%leading to more efficient and context-aware movement.
%Our method categorizes spaces into different room types based on the information of the object nodes and structural characteristics. This categorization allows for more tailored navigation strategies within each room type.
		
\subsection{Data structure of 3DSGs}
%Visualization and data format of the 3DSGs

To leverage the capabilities of LLMs for down-streaming navigation tasks, 
3DSGs are represented as a NetworkX graph object~\cite{HSS08}, \cite{Rana2023SayPlanGL}
in a JSON data format in a text-serialized manner as depicted in Figure~\ref{fig:3DSG_illu}. 
%JSON is a lightweight data interchange format that is easy for humans to read and write and easy for machines to parse and generate. 
By converting the 3DSG into JSON, it becomes a structured input that can be directly utilized by LLMs 
for various AI tasks, such as semantic search and planning for robotic navigation~\cite{Rana2023SayPlanGL}.

%+ 3DSG visualization 
%
%{\color{red}{an example of the node structure}}

%Key new points and results 
%
%  **Node attributes**
%
%    Our research introduces a novel framework for defining node attributes in 3DSGs. Each node is enriched with a comprehensive set of attributes that include not only geometric properties such as volume and surface area but also semantic descriptors like object function and room purpose, derived through LLM analysis.
%
%    
%
%  **Room types - + polling**
%
%    Our method categorizes spaces into different room types based on **the information of the object nodes and structural characteristics**, allowing for more tailored navigation strategies within each room type.
%
%    We implement a polling mechanism that leverages the LLM's capabilities to evaluate and select the most reasonable room type.
%
%      different polling strategies 
%
%      reducing the requirement on the LLM itself 
%
%      combating the issue with hallucination and reducing uncertainty 

%% file: res.tex
%The results of our experiments demonstrate that the proposed system significantly outperforms traditional navigation methods in terms of accuracy, efficiency, and adaptability. The hierarchical 3DSGs, coupled with the semantic understanding provided by LLMs, enable robots to navigate complex environments with a level of intelligence and precision that was previously unattainable.
%
%{\color{red}{These tasks are assessed in two expansive environments, including a large office floor spanning 37 rooms and 150 interactable assets and objects, and a three-storey house with 28 rooms and 112 objects.}}
%
%We use the ERNIE 3.5 provided by Baidu for the numerical experiments presented in this section. 
%The temperature is set to 0 such that the response obtained from the LLM is rather deterministic. 
%

%The primary objective of our experiments is to validate 
%the effectiveness of the proposed hierarchical 3D Semantic Graphs (3DSGs) system, 
%which integrates Large Language Models (LLMs) for intelligent spatial perception in indoor navigation scenarios. 
%%We aim to demonstrate the system's superiority over traditional navigation methods in terms of accuracy, efficiency, and adaptability.
%%The system implementation is based on the methodological framework detailed in the paper. 
%We utilize the ERNIE 3.5 model provided by Baidu for the numerical experiments. 
%The temperature parameter for the LLM is set to 0, ensuring stable responses from the model. 
The primary goal of our experiments is to validate the effectiveness of the proposed 
method of generating hierarchical 3DSGs. 
%which integrates LLMs for intelligent spatial perception in indoor navigation scenarios. 
If not stated otherwise, the ERNIE 3.5 model provided by Baidu is utilized for the numerical experiments, 
with the temperature parameter set to 0.1 to ensure stable responses.

\subsection{3DSG generation}

\begin{figure}[!h]
%         \vspace{-0.6cm}   %调整图片与上文的垂直距离  
%	\setlength{\abovecaptionskip}{0.cm} %调整标题上方的距离   
%	\setlength{\abovecaptionskip}{0.1cm} %调整标题下方的距离 	  
   % \ContinuedFloat
    \centering
%    \subfigure[Illustration of the indoor tour-guide robot (left) and 
%    the indoor delivery robot (right)]{
        \includegraphics[width=0.37\textwidth]{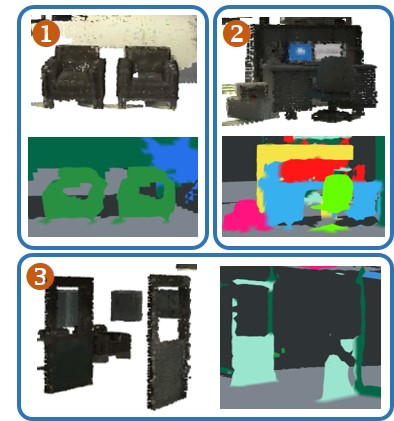}
%        }
        \caption{Examples and comparison of point-cloud and mesh representations of object nodes}
    \label{fig:pc_mesh_comp}
\end{figure}

%uHumans dataset 
%
%Point to Figure 1 and briefly describe the layers 
%
%The hierarchical 3DSGs are constructed in real-time as the robot navigates its surroundings, 
%with each layer serving a specific purpose in the overall navigation process.
%
%Data is collected using a combination of RGB-D cameras and LiDAR sensors, 
%which provide the necessary geometric and semantic information for constructing the 3DSGs.
%The 2D semantic segmentation network YOLOv8 is employed for pixel-level semantic segmentation of the RGB images. 
%Depth information is either directly provided by the RGB-D camera or retrieved through stereo matching for stereo cameras.
We illustrate the 3DSG of the office scene 
of the uHumans2 dataset~\cite{Rosinol2021KimeraFS} generated with the proposed method 
already in Figure~\ref{fig:3DSG_illu}. 
The point cloud of the object nodes is not shown in this whole view of the 3DSG 
for clarity of the visualization. 
Figure~\ref{fig:pc_mesh_comp} presents a few examples 
and a comparison with their counterparts taking the form of a mesh as in the Hydra framework~\cite{HCH24}. 
It can be seen that this point-cloud form 
serves as a more precise representation of the object nodes, 
which is essential for guaranteeing a satisfactory robot navigation and manipulation performance. 
 
\begin{figure}[!h]

   % \ContinuedFloat
    \centering
%    \subfigure[Illustration of the indoor tour-guide robot (left) and 
%    the indoor delivery robot (right)]{
        \includegraphics[width=0.48\textwidth]{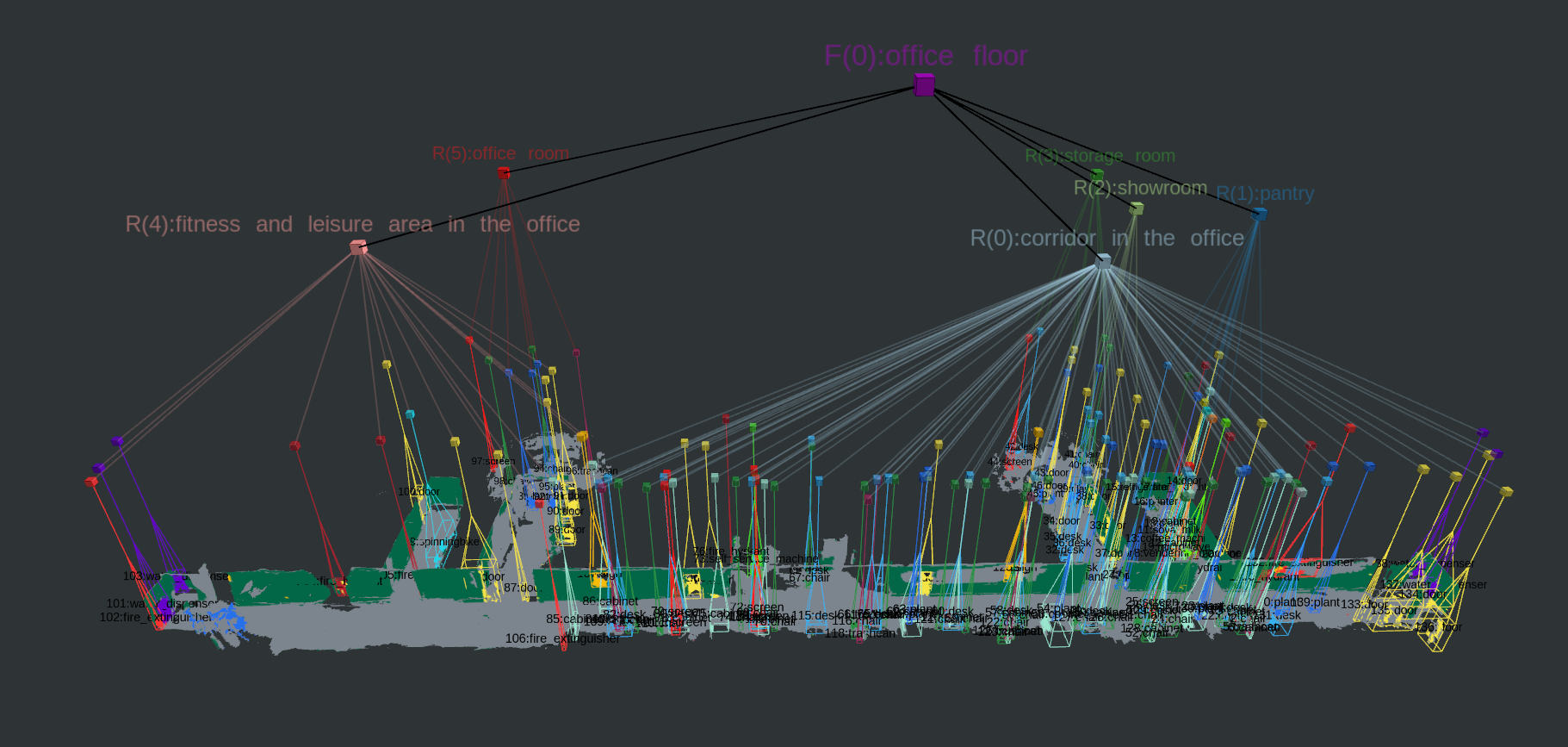}
%        }
        \caption{3DSG of an office scenario constructed in real-time with a wheeled robot}
    \label{fig:3DSG_S02}
\end{figure}

In addition to testing with datasets, we have also constructed 3DSGs 
in real-time with a wheeled robot developed in our company for indoor scenarios. 
Data is collected using a combination of visual and LiDAR sensors, 
providing the necessary geometric and semantic information for constructing the 3DSGs. 
In particular, we use the ORBBEC Femto Bolt RGB-D camera to obtain RGB and depth images of the scenario. 
Coupling a 2D LiDAR sensor, a wheel odometry, and an inertial measurement unit (IMU)~\cite{Labb2018RTABMapAA} 
equipped on the robot device 
leads to pose estimates essential for the generation of {\textit{Layer-1}}. 
%Depth information is either directly provided by the RGB-D camera or retrieved through stereo matching for stereo cameras.
The 2D semantic segmentation network YOLOv8 is employed for pixel-level semantic segmentation of the RGB images.  
Figure~\ref{fig:3DSG_S02} presents the resulting 3DSG created for an office scene of our company. 
It consists of a fundamental layer, 
an object layer, a room layer, and a floor layer with nodes annotated. 
\begin{figure*}[htbp]
   % \ContinuedFloat
    \centering
%    \subfigure[Illustration of the indoor tour-guide robot (left) and 
%    the indoor delivery robot (right)]{
        \includegraphics[width=0.8\textwidth]{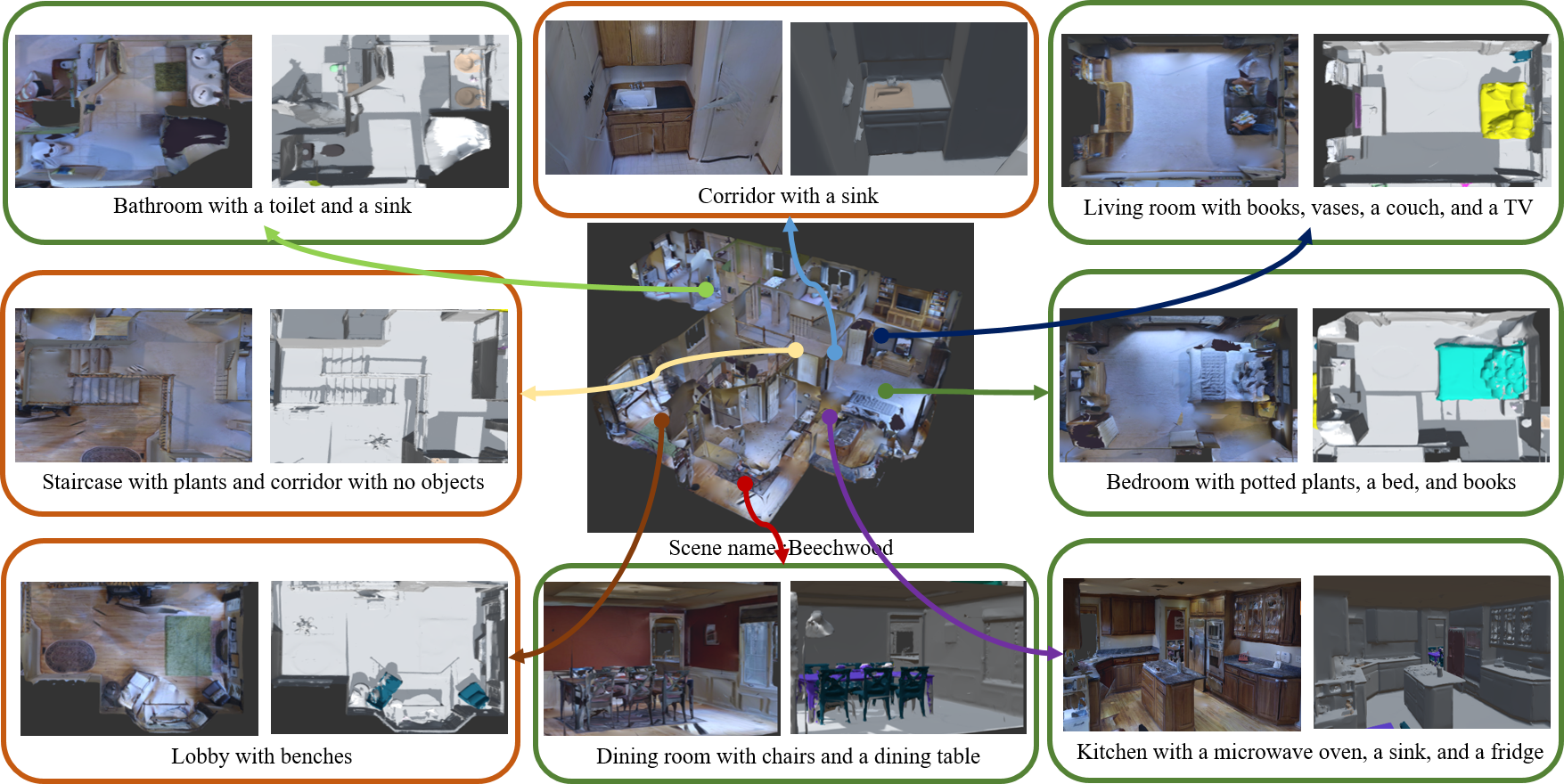}
%        }
        \caption{A scene in the Stanford 3D Scene Graph dataset~\cite{Armeni20193DSG} where 
        the point cloud representation, semantic mesh, and object information of various room segments are presented}
    \label{fig:Beechwood_iilu}
\end{figure*}
In summary, these results indicate that with the proposed scheme, 3DSGs can be built 
 using not only existing datasets but also on-board sensors of a robotic device in real-time. 

Although preliminary results of applying the 3DSG shown in Figure~\ref{fig:3DSG_S02} 
for robot navigation in the corresponding office scene are not shown due to the page limitation, 
the performance is satisfactory. 
On the other hand, it should be noted that there are still not any standardized or widely acknowledged 
evaluation criteria and metrics for 3DSGs. 
Generally speaking, the geometric accuracy of the fundamental layer is decided by 
the performance of the pose estimation and the mesh generation schemes
which in our case have been thoroughly evaluated in~\cite{Labb2018RTABMapAA} and~\cite{RA20}, respectively. 
In the sequel, we focus on the semantic aspect of the 3DSGs and 
assess the performance of the proposed spatial perception approach. 

\subsection{Node description}
The experiments shown in this section 
are mainly conducted with 
the Stanford 3D Scene Graph dataset ``Klickitat''~\cite{Armeni20193DSG} 
corresponding to an 
expansive and complex indoor scenario of 
a three-floor residential building with 28 rooms and 112 objects. 
A 3DSG is available, whereas the raw data can be extracted using the iGibson environment~\cite{shenigibson} 
and used to supplement desired information to the 3DSG. 
%Its complexity and variability, 
%providing a robust testbed for evaluating the system's performance.
%The hierarchical 3DSGs are constructed in real-time as the robot navigates its surroundings, with each layer serving a specific purpose in the overall navigation process.
%In the sequel, 
The correctness, efficiency, and stability of 
the semantic search and the planning~\cite{Rana2023SayPlanGL} 
are treated as evaluation criteria in these experiments. 

\begin{figure}[!h]
%         \vspace{-0.8cm}   %调整图片与上文的垂直距离  
%	\setlength{\abovecaptionskip}{0.cm} %调整标题上方的距离   
%	\setlength{\abovecaptionskip}{0.1cm} %调整标题下方的距离 	 
   % \ContinuedFloat
    \centering
%    \subfigure[Illustration of the indoor tour-guide robot (left) and 
%    the indoor delivery robot (right)]{
        \includegraphics[width=0.48\textwidth]{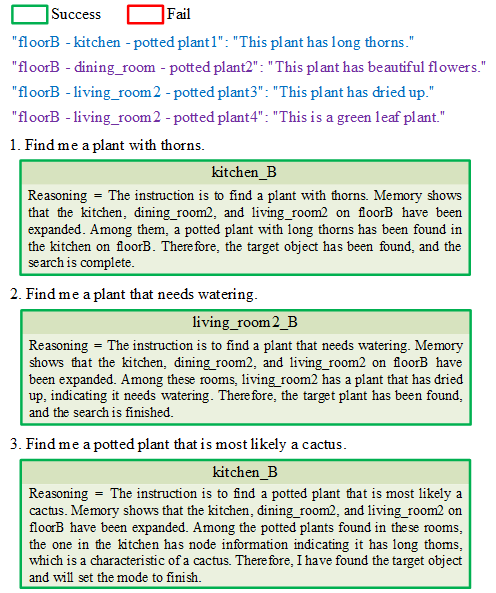}
%        }
        \caption{Results of semantic search in the case of different task queries 
        when using 3DSG with and without detailed object node descriptions, respectively}
    \label{fig:res_nodes}
%    \vspace{-0.7cm}   %调整图片与上文的垂直距离  
\end{figure}

\begin{figure}[!h]
   % \ContinuedFloat
    \centering
%    \subfigure[Illustration of the indoor tour-guide robot (left) and 
%    the indoor delivery robot (right)]{
        \includegraphics[width=0.48\textwidth]{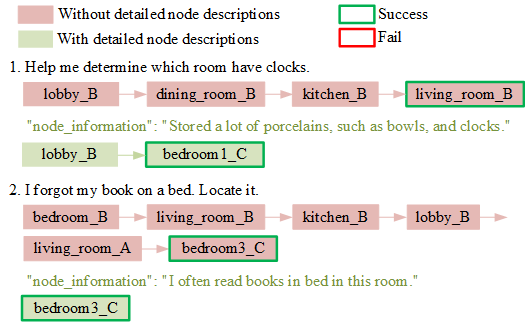}
%        }
        \caption{Results of semantic search in the case of different task queries 
        when using 3DSG with and without detailed room node descriptions, respectively}
    \label{fig:res_room_nodes}
\end{figure}

Figure~\ref{fig:res_nodes} and Figure~\ref{fig:res_room_nodes} 
illustrate the results of semantic search in the case of different task queries 
when using 3DSG with and without detailed descriptions 
for object nodes of {\textit{Layer-2}} and room nodes of {\textit{Layer-3}}, respectively. 
The results demonstrate the enhanced semantic search efficiency 
when detailed node descriptions are included.

\subsection{Room classification}
\label{subsec:res_room}
This section presents the findings from the experiments with the room classification as a focus conducted 
using various LLMs and schemes developed along the investigation. 
For these experiments, we use the 35 scenarios of family apartment buildings in 
the Stanford 3D Scene Graph dataset~\cite{Armeni20193DSG} 
comprising 483 rooms and 2339 objects categorized in 35 types. 
The reason why we have turned to this dataset is twofold: first, 
it is a very large dataset encompassing representative 
and complex indoor household scenarios (cf. Figure~\ref{fig:Beechwood_iilu} for an example); 
second, ground truth room labels are available in this dataset and can be taken as a reference for further evaluation. 
Hence, compared to the 3DSGs we have constructed in real-time in relatively limited scenarios, 
we believe using the dataset can better shed light on future development of room classification approaches based on LLMs.

%
%The dataset, part of the 3D Scene Graph, includes multi-layered graph data with a focus on room and object layers. 
%The analysis was restricted to rooms that contained objects, as object information is pivotal for room classification.

\begin{figure}[!h]

   % \ContinuedFloat
    \centering
        
     \subfigure[Querying the LLMs directly with information
of the objects contained in the room node]{
        \includegraphics[width=0.44\textwidth]{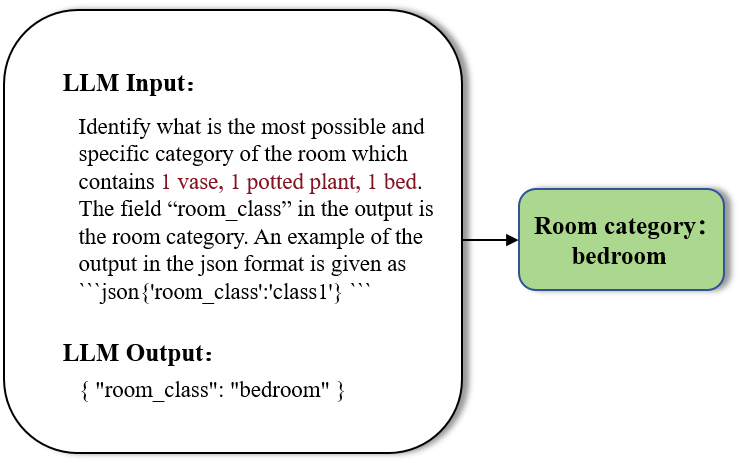}
        \label{fig:prompt_template_ori}}

     \subfigure[With the polling strategy incorporated]{
        \includegraphics[width=0.44\textwidth]{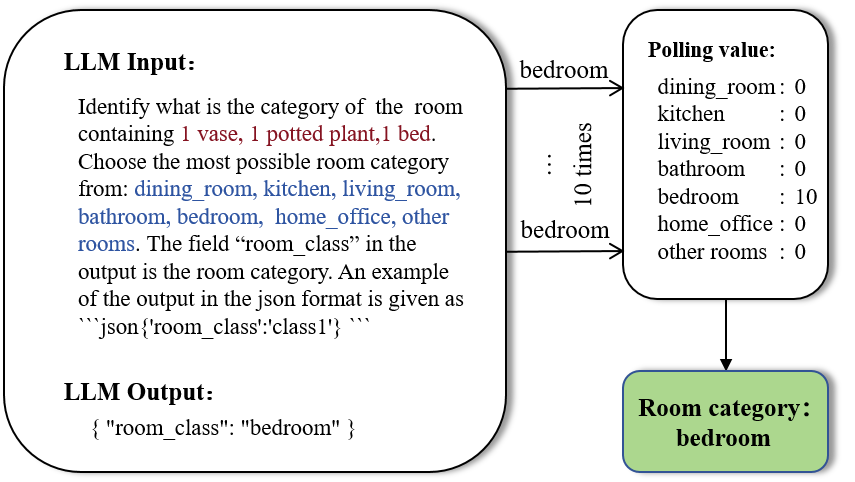}
        \label{fig:prompt_template_polling}
        }
%        \caption{Results of semantic search and planning in the case of two task queries with negation}
%\end{figure}        
%      \begin{figure}[!h]  
%         \vspace{-0.4cm}   %调整图片与上文的垂直距离  
%	\setlength{\abovecaptionskip}{0.cm} %调整标题上方的距离   
%	\setlength{\abovecaptionskip}{0.cm} %调整标题下方的距离 	   
%	\setlength{\belowdisplayskip}{3pt} 	
        \caption{Illustration of two versions of the LLM-based room classification method and the prompts designed, 
        with the node information that varies from node to node and the set of typical room labels that can be adjusted 
        marked in a dark red and a light blue color, respectively}
    \label{fig:prompt_template}
\end{figure}

The initial approach involves querying the LLMs directly with information
of the objects contained in the room node to identify the room category 
(cf. Figure~\ref{fig:prompt_template_ori} for an example). 
As presented in Table~\ref{tab:acc}, three distinct LLMs are tested: ERNIE-Speed-8K, KIMI-8K, and ERNIE-3.5-8K. 
The ERNIE-3.5-8K model, developed by Baidu Qianfan, demonstrates the highest accuracy, 
thus having been used for the subsequent analysis.
\begin{table}[h!]
  \begin{center}
    \caption{Accuracy of room classification when various LLMs and strategies are used}
    \begin{tabular}{ccc} % <-- Alignments: 1st column left, 2nd middle and 3rd right, with vertical lines in between\\
      \hline
      \textbf{mode/method} & \textbf{accuracy} & \textbf{number of rooms annotated} \\
      \hline
      ERNIE-Speed-8K & 0.7060 & 483 \\
      KIMI-8K & 0.7721 & 483 \\
      ERNIE-3.5-8K & 0.8178 & 483\\
      ERNIE-3.5-8K + polling & 0.9527 & 358 \\
      \hline
    \end{tabular}
    \label{tab:acc}
  \end{center}
\end{table}
%      FBTO & 0.9274 & 358 \\
%%%%%%%%%%%%%
% findings of initial experiments and analyzing indoor household scenarios combined with common-sense 
% 
% not all segments of an indoor scenario can be well described with a single-word label
% 
% some can be multi-functional
% 
% strategy: do not annotate all rooms with a single label, but guarantee that once a room label is given, it should be accurate 
% 
% multi-functional segments are captioned with a concise description 
% 
% usually only room labels are provided for the semantic search and task planning  
% 
% if the room label field is empty, node description is used 
% 
% further room labels could be obtained via human-device interaction 

In these initial experiments and analysis of these indoor household scenarios, 
we have checked the rooms for which the labels inferred by the LLM deviate 
from the ground truth given in the dataset~\cite{Armeni20193DSG}, 
and we notice that not all segments of an indoor scenario can be well described with a single-word label, 
while some can be multi-functional. 
In addition, it is important to note that 
the rate of classification accuracy only measures the discrepancy between the estimated label and  
the ground truth label which technically cannot be regarded as absolutely right, 
for the label of a segment in an indoor scenario is not definite and can also be subjective. 
Therefore, our strategy is to not annotate all rooms with a single label, 
but to ensure that once a room label is given, it should be accurate. 
Multi-functional segments are captioned with a concise description. 
Typically, only room labels are provided for semantic search and task planning. 
If the room label field is empty, the node description is used instead. 
%Further room labels could be obtained via human-device interaction. 
%When it is possible, annotating a room node with an accurate label is important, 
%for the token size can be reduced and providing the LLM with 
%concise and explicit information prevents it from hallucinating effectively. 

Further investigation into the impact of object categories on room classification 
reveals significant disparities in accuracy across different object categories. 
\begin{figure}[!h]
   % \ContinuedFloat
    \centering
%    \subfigure[Illustration of the indoor tour-guide robot (left) and 
%    the indoor delivery robot (right)]{
        \includegraphics[width=0.48\textwidth]{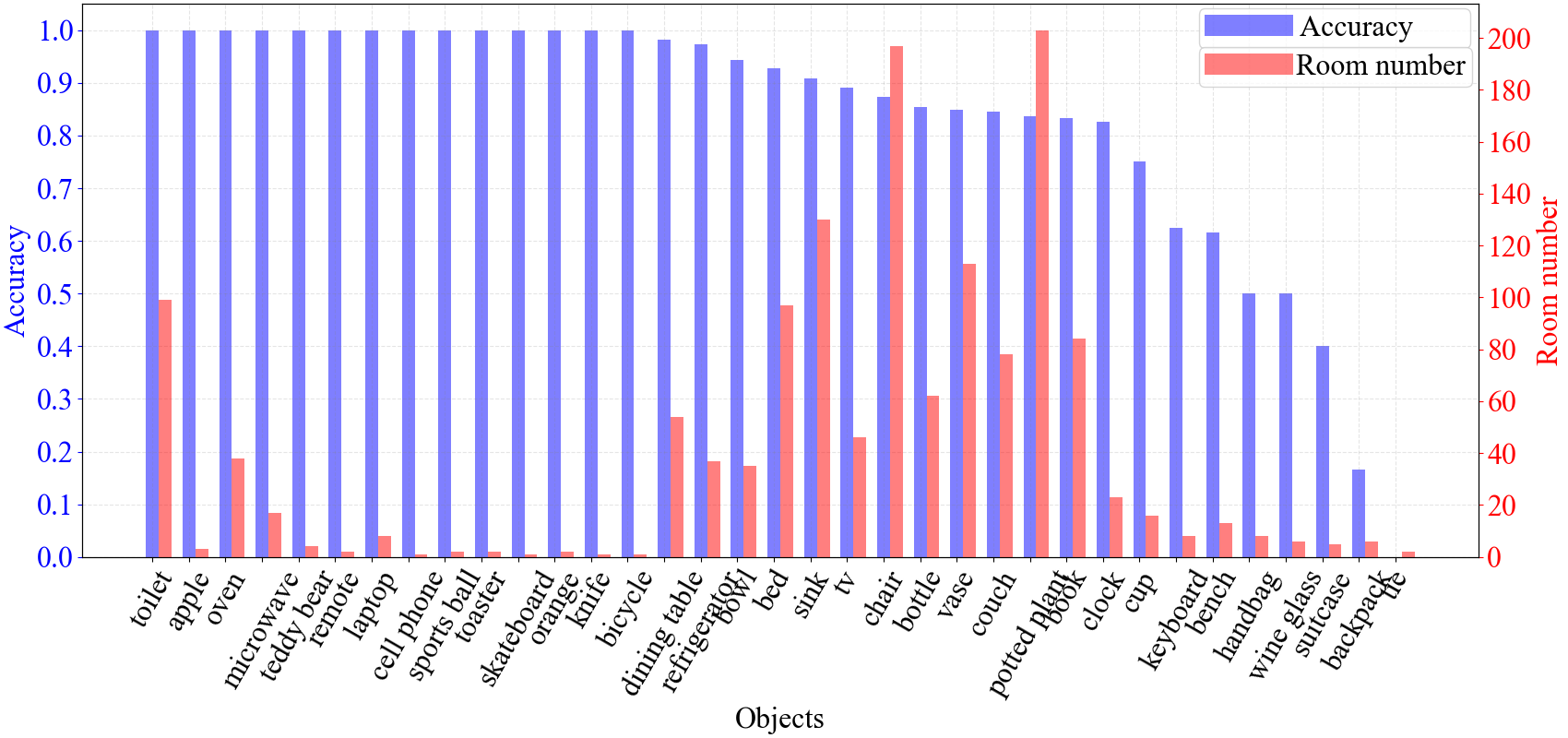}
%        }
        \caption{Illustration of the impact of object categories on the accuracy of the room classification}
    \label{fig:res_object_analysis}
\end{figure}
As presented in Figure~\ref{fig:res_object_analysis}, rooms containing specific objects, such as ``toilet'', 
show a perfect classification accuracy, implying that 
these objects can be treated as reliable indicators for room categorization. 
%By filtering rooms based on the presence of such typical objects (Filtered by Typical Objects, FBTO), 
%the classification accuracy is improved to 0.9247.

\begin{figure}[!h]
   % \ContinuedFloat
    \centering
%    \subfigure[Illustration of the indoor tour-guide robot (left) and 
%    the indoor delivery robot (right)]{
        \includegraphics[width=0.48\textwidth]{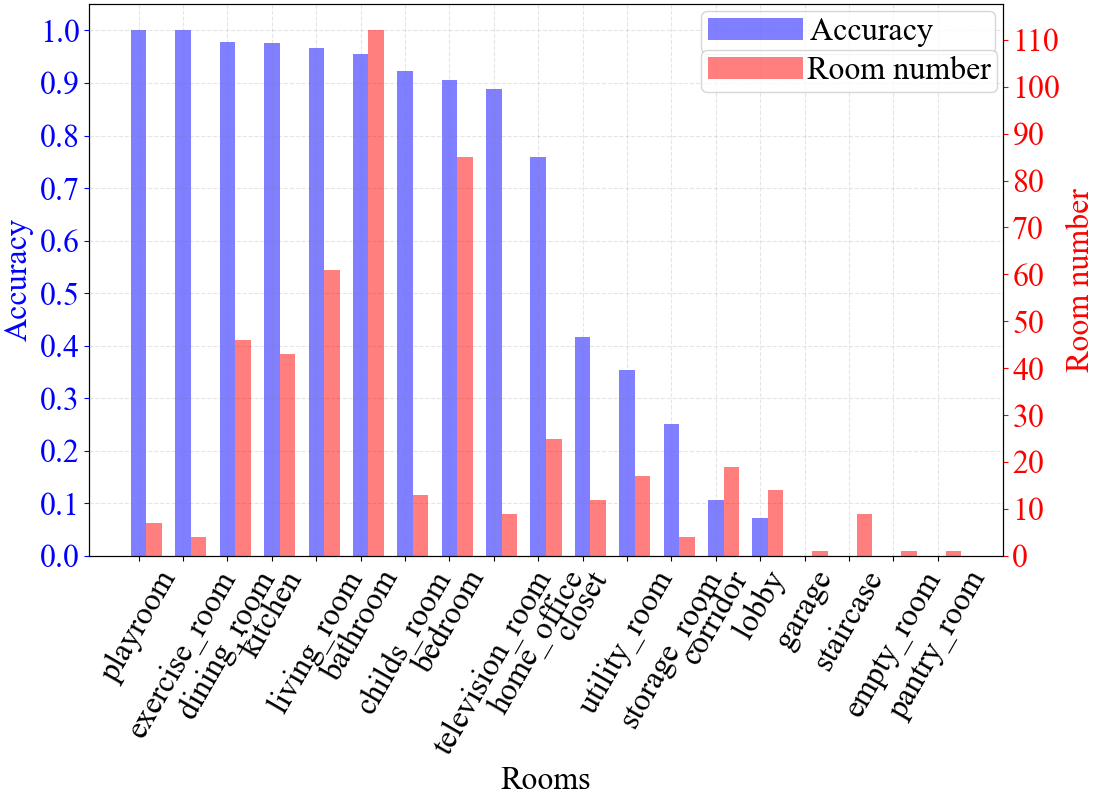}
%        }
        \caption{Accuracy of the room classification with respect to room categories}
    \label{fig:res_room_analysis}
\end{figure}

The relationship between room categories and classification accuracy is also analyzed. 
Figure~\ref{fig:res_room_analysis} shows that for 
certain room types, such as ``dining$\_$room'', the classification accuracy is high. 
%Identifying these rooms as typical and using them as references significantly contributed to the reliability of room categorization.
These rooms are then identified and used as the typical room labels in the polling approach 
introduced in Section~\ref{subsec:3DSG}. 
An example is illustrated in Figure~\ref{fig:prompt_template_polling}. 
%The polling method was employed to refine room classification by considering room categories as voting options 
%as illustrated in Figure~\ref{fig:prompt_template}. 
% based on the observations and findings regarding the typical rooms and typical objects, 
% 
%The 'LLM+polling+top1' method achieved an accuracy of 0.9497. 
The resulting enhanced accuracy of classification is presented in Table~\ref{tab:acc}. 
Only when the polling result is a ``full-score'' as illustrated in Figure~\ref{fig:prompt_template_polling}, 
the room is captioned with the selected label. 
This strategy helps identify rooms with diverse object nodes and multiple functions 
for which a concise node description is more suitable and effective for subsequent robotic navigation tasks 
compared to a single room label.  
% corridor 
A few examples of such cases are shown in Table~\ref{tab:polling}. 
Therefore, the polling strategy proposed in this paper not only gives rise to 
an enhanced credibility of the resulting room label but also 
leads to an effective scheme of identifying and screening multi-functional segments of an indoor scenario. 
This intelligent ``hybrid'' room annotation method fully exploits the benefits of using LLMs 
and is tailored for complex indoor scenarios. 
\begin{table}[h!]
  \begin{center}
    \caption{Examples of the polling results for multi-functional room segments}
    \begin{tabularx}{0.49\textwidth}{XXX} 
      \hline
%      \makebox[0.01\textwidth][c]{name} & \makebox[0.01\textwidth][c]{taskA} & \makebox[0.2\textwidth][c]{taskB} \\
      \textbf{baseline} & \textbf{list of objects} & \textbf{polling results} \\
      \hline
      home$\_$office & 1 bowl, 3 chairs, 3 potted plants, 1 laptop, 5 books, 1 clock, 3 vases & living$\_$room:~5$|$ \newline bedroom:~5\\
      \hline
      bedroom & 1 bench, 1 chair, 2 couches, 3 potted plants, 1 bed & living$\_$room:~5$|$ \newline bedroom:~5 \\
      \hline
      utility$\_$room &1 backpack, 3 bottles, 2 cups, 1 chair, 4 potted plant, 1 sink & kitchen:~5$|$bathroom:~3$|$ \newline other room:~2 \\
      \hline
      corridor & 2 bowls, 1 vase & dining$\_$room:~7$|$ \newline living$\_$room:~2$|$kitchen:~1 \\
      \hline
    \end{tabularx}
    \label{tab:polling}
  \end{center}
%  \vspace{-0.5cm}   %调整图片与上文的垂直距离  
\end{table}